\documentclass[conference]{IEEEtran}
\IEEEoverridecommandlockouts    

\usepackage{multirow}
\usepackage[ruled,vlined]{algorithm2e}
\usepackage{cite}
\usepackage{amsmath,amssymb,amsfonts}
\usepackage{algorithmic}
\usepackage{graphicx}
\usepackage{textcomp}
\usepackage{booktabs}
\usepackage{hyperref}
\usepackage{xcolor}
\usepackage{tcolorbox}
\usepackage{amssymb}

\usepackage{listings}
\tcbuselibrary{skins,breakable}
\definecolor{codebg}{RGB}{248,248,248}
\definecolor{dangerred}{RGB}{220,53,69}
\definecolor{successgreen}{RGB}{40,167,69}
\definecolor{lightgray}{RGB}{200,200,200}

\newtcolorbox{badcode}{
   colback=codebg,
   colframe=lightgray,
   boxrule=0.5pt,
   left=2mm,
   right=2mm,
   top=5mm,
   bottom=0mm,
   before skip=4pt,
   after skip=4pt,
   enhanced,
   overlay={\node[anchor=north west,outer sep=5pt,inner sep=3pt] at (frame.north west) {\textcolor{dangerred}{\textbf{$\times$ VULNERABLE}}};}
}

\newtcolorbox{goodcode}{
   colback=codebg,
   colframe=lightgray,
   boxrule=0.5pt,
   left=2mm,
   right=2mm,
   top=5mm,
   bottom=0mm,
   before skip=4pt,
   after skip=4pt,
   enhanced,
   overlay={\node[anchor=north west,outer sep=5pt,inner sep=3pt] at (frame.north west) {\textcolor{successgreen}{\textbf{$\checkmark$ SECURE}}};}
}

\title{From Leaderboard to Deployment: Code Quality Challenges in AV Perception Repositories}

\author{
    Mateus Karvat Camara$^{1}$, Bram Adams$^{1}$, and Sidney Givigi$^{1}$
        \thanks{$^{1}$School of Computing, Queen’s University. Kingston, ON, Canada. {\tt\small mateus.karvat@queensu.ca, bram.adams@queensu.ca, sidney.givigi@queensu.ca}}
}

\begin{document}
	
\maketitle

\begin{abstract}
    Autonomous vehicle (AV) perception models are typically evaluated solely on benchmark performance metrics, with limited attention to code quality, production readiness and long-term maintainability. This creates a significant gap between research excellence and real-world deployment in safety-critical systems subject to international safety standards. To address this gap, we present the first large-scale empirical study of software quality in AV perception repositories, systematically analyzing 178 unique models from the KITTI and NuScenes 3D Object Detection leaderboards. Using static analysis tools (Pylint, Bandit, and Radon), we evaluated code errors, security vulnerabilities, maintainability, and development practices. Our findings revealed that only $7.3\%$ of the studied repositories meet basic production-readiness criteria, defined as having zero critical errors and no high-severity security vulnerabilities. Security issues are highly concentrated, with the top five issues responsible for almost $80\%$ of occurrences, which prompted us to develop a set of actionable guidelines to prevent them. Additionally, the adoption of Continuous Integration/Continuous Deployment pipelines was correlated with better code maintainability. Our findings highlight that leaderboard performance does not reflect production readiness and that targeted interventions could substantially improve the quality and safety of AV perception code.
\end{abstract}

\begin{IEEEkeywords}
    Autonomous vehicles, perception models, autonomous vehicle safety, software security, software quality.
\end{IEEEkeywords}

\section{Introduction}
Perception models are key components within the pipeline of Autonomous Vehicles (AVs), enabling them to interpret their surroundings and make safety-critical decisions. While research in AV perception has advanced rapidly, with models achieving impressive performance on benchmark leaderboards such as KITTI~\cite{geiger_2013_kitti} and NuScenes~\cite{caesar_2019_nuscenes}, the transition from research code to production-ready systems remains challenging. Production deployment requires not only high accuracy, but also code that is maintainable, secure, and free of critical errors that could compromise safety in real-world scenarios.

State-of-the-art perception models are typically evaluated based on their detection accuracy, with leaderboard rankings serving as one of the primary measures of success. However, these rankings provide no assessment of code quality: a top-performing model may have excellent detection metrics while having a codebase that is undocumented, filled with security vulnerabilities, or prone to critical runtime errors. As such, a significant gap exists between research excellence and production readiness, creating barriers to adapting high-performing research models for production use, often requiring reimplementation from scratch.

Despite the importance of code quality in AV systems, no large-scale empirical study has systematically analyzed the deployment readiness of AV perception repositories.  Given that AVs are safety-critical systems that must meet stringent quality standards~\cite{MISRA2012,iso26262,iso21448}, understanding the current state of code quality in perception models is essential for bridging the research-production gap.

Motivated by this, we conducted the first comprehensive software quality analysis of AV perception models, analyzing $178$ unique repositories from the KITTI and NuScenes 3D Object Detection leaderboards. Using the static analysis tools Pylint~\cite{pylint_2025}, Bandit~\cite{bandit_2025}, and Radon~\cite{radon_2025}, we evaluated each repository across multiple quality dimensions: code errors, security vulnerabilities, maintainability, and development practices. We investigated relationships between code metrics and team characteristics to understand what factors contribute to higher quality codebases, and analyzed the most common issues to develop guidelines for their prevention.

Our analysis revealed significant quality challenges in the current state of AV perception code. Only $7.3\%$ of repositories meet production-readiness, defined as having zero critical errors and no high-severity security vulnerabilities. Moreover, security issues exhibit a strong concentration, with the top five issues accounting for almost $80\%$ of all occurrences.

Our main contributions are: 

\begin{itemize}
	\item the first large-scale empirical study of software quality in AV perception models, analyzing $178$ repositories from the two largest 3D Object Detection leaderboards.
	\item the development of guidelines targeting the most common security issues, which account for $80\%$ of occurrences.
	\item insights on development practices showing that Continuous Integration/Continuous Deployment (CI/CD) adoption correlates with higher code quality, despite rarely being used in the research community.
\end{itemize}

Our findings highlight the substantial gap between research excellence and production readiness in AV perception models. By identifying common quality issues and their root causes, we provide actionable guidance for researchers and practitioners working to bridge this gap and develop perception models that are suitable for deployment in safety-critical AV systems.

\section{Related Work}
\label{sec:related_work}

The requirements for the safe deployment of safety-critical automotive software have been established by safety standards such as ISO 26262~\cite{iso26262}, which mandates the use of coding guidelines, static analysis, and systematic testing. However, it was designed for deterministic systems and does not fully address challenges introduced by Machine Learning components~\cite{salay_2018_analysis}. ISO 21448 (SOTIF)~\cite{iso21448} was introduced to complement it by addressing safety risks arising from functional insufficiencies in perception and decision-making, rather than hardware or software malfunctions. Together, these standards establish that software quality in AV perception systems is not merely a development convenience but a safety requirement.

Koopman and Wagner~\cite{koopman_2016_challenges} argued that system-level testing alone is insufficient to ensure AV safety, and that rigorous software development processes are essential. This concern is present in the work of Tabani et al.~\cite{tabani_2019_assessing}, who assessed the adherence of Baidu Apollo's autonomous driving framework to ISO 26262 software guidelines. Their analysis revealed significant compliance gaps, particularly in modules relying on GPU-accelerated deep learning libraries that were not designed for safety-critical applications, highlighting a broader problem: AV software is frequently built on components not designed for the automotive domain. While their study focused on a single production framework, our work addresses the research repositories from which such systems are often derived. 

More recently, Cheng et al.~\cite{cheng_2025_unveiling} used CodeQL to analyze security vulnerabilities in repositories for autonomous driving platforms, finding a direct link between vulnerability severity and system performance. Kochanthara et al.~\cite{kochanthara_2024_safety} derived safety requirements for AV perception systems using ISO 26262 and ISO 21448, showing that ML-based perception pipelines introduce specific risks that require targeted mitigation strategies. 

These studies confirm that code-level quality issues in perception systems have concrete safety implications, yet neither provides a large-scale quantitative assessment of the perception model research repositories that often serve as starting points for production implementations. To the best of our knowledge, no prior study has systematically analyzed the code quality of publicly available AV perception model repositories at scale, examining errors, security vulnerabilities, and maintainability, which is the gap this work addresses.

\section{Dataset}
\label{sec:dataset}

We selected repositories from the KITTI~\cite{geiger_2013_kitti} and NuScenes~\cite{caesar_2019_nuscenes} 3D Object Detection leaderboards. Since the KITTI leaderboard is subdivided into three object categories (Car, Pedestrian, and Cyclist) with distinct rankings, we considered all models listed in any of the categories, disregarding models appearing in multiple categories. Our data collection was conducted in November 2025, and resulted in a starting pool of $389$ models from the KITTI leaderboard and $330$ models from the NuScenes leaderboard.

We then filtered this pool, first removing models that did not have a repository link, with $164$ models from KITTI remaining and $134$ from NuScenes. Given that these are the largest 3D Object Detection leaderboards for AVs, we found it surprising that $421$ models did not have any repository link. That amounts to almost $60\%$ of the models that cannot have their results easily replicated and whose reuse is severely hindered.

The two pools of models were merged, and two duplicates appearing on both leaderboards were removed. We then verified all repositories using our data scraping script, excluding $11$ models due to invalid links (404 errors or deleted repositories). Next, we removed $72$ model variants (e.g., UVTR-Camera and UVTR-LiDAR) since model variations share the same codebase. After that, we removed four repositories that had valid links but with inaccessible data, either because the repository was empty, archived, or private.

After these steps, $209$ repositories remained. We then assessed their size based on their Python files. Despite removing roughly $70\%$ of our initial pool, $13.4\%$ of the remaining repositories had no Python code and consisted exclusively of markdown files (used as README/documentation on GitHub). Consequently, $28$ repositories were removed for containing solely markdown files, and three additional repositories were removed due to duplicate links or invalid filenames.

Our final dataset comprised $178$ repositories, whose size ranged from $600$ to $184.9$k Source Lines of Code (SLOC) (mean $23.1$k, median $14$k, IQR: $7.4$k--$23.3$k).

\section{Methodology}
\label{sec:methodology}

To evaluate code quality and security in AV perception repositories, we employed the static analysis tools Pylint~\cite{pylint_2025} (v2.16.2), Bandit~\cite{bandit_2025} (v1.8.6), and Radon~\cite{radon_2025} (v6.0.1). 

Pylint is a popular Python linter, a tool that analyzes code to identify errors, enforce coding standards, and provide refactoring suggestions. We used it to identify code errors in the repositories, with a special focus on those that prevent code execution or cause runtime crashes, which we defined as critical errors. These included import errors, name errors, syntax errors, type errors, and logic errors.

Bandit was used to identify security issues, which are critical for production software, yet often overlooked in research code. For each repository, we collected the total number of security issues and the number classified as high-severity. We also analyzed Bandit's reports to extract representative examples of each issue type and collect severity classifications.

Radon was used to evaluate the total SLOC of each repository, as well as the maintainability index (MI)~\cite{coleman_1994_using}, a metric that indicates how easy it is to maintain and change code.

Since our dataset exhibited non-normal distributions and contained outliers, we employed Spearman correlation coefficients to quantify relationships between continuous variables due to its robustness to this type of data. To compare distributions between independent groups, we used the Mann-Whitney U test ($\alpha=0.05$), a non-parametric alternative to the independent t-test that does not assume normality.

Finally, to better understand how team characteristics and development practices influence code quality, we used the GitHub API to collect repository metrics, contributor counts, issue statistics, and popularity metrics (stars and forks). We also assessed development practices such as CI/CD pipeline adoption (GitHub actions or Travis CI) and test infrastructure, as these represent systematic quality assurance mechanisms that could influence error detection and code maintainability.

\section{Results}
\label{sec:results}

\subsection{Production Readiness}
\label{sec:res_production_readiness}

Given that AVs are safety-critical systems, we were inspired by software guidelines for critical systems~\cite{MISRA2012} and ML production-readiness frameworks~\cite{breck_2017_mltest} to define production-readiness criteria, which we considered as zero critical errors and no high-severity security issues. Under these criteria, \textbf{only $7.3\%$ of the repositories qualify as production-ready} ($13$ out of $178$). Most production-ready repositories fall below our dataset's median SLOC ($14$k), with only three exceptions ($14.5$k, $17.9$k, and $51.8$k).

Regarding errors, $97.2\%$ of the repositories exhibited at least one, with only five being error-free. The median number of errors was $29$ (mean: $55.7$), ranging from $0$ to $1,263$ (IQR: $10\text{--}52$). Comparing repository sizes, we found error-free repositories to be considerably smaller. While the overall mean SLOC in our dataset was $23.1$k, error-free repositories averaged merely $5$k SLOC, compared to $23.6$k SLOC for repositories with errors.

\begin{figure}
\centering
    \includegraphics[width=.88\columnwidth]{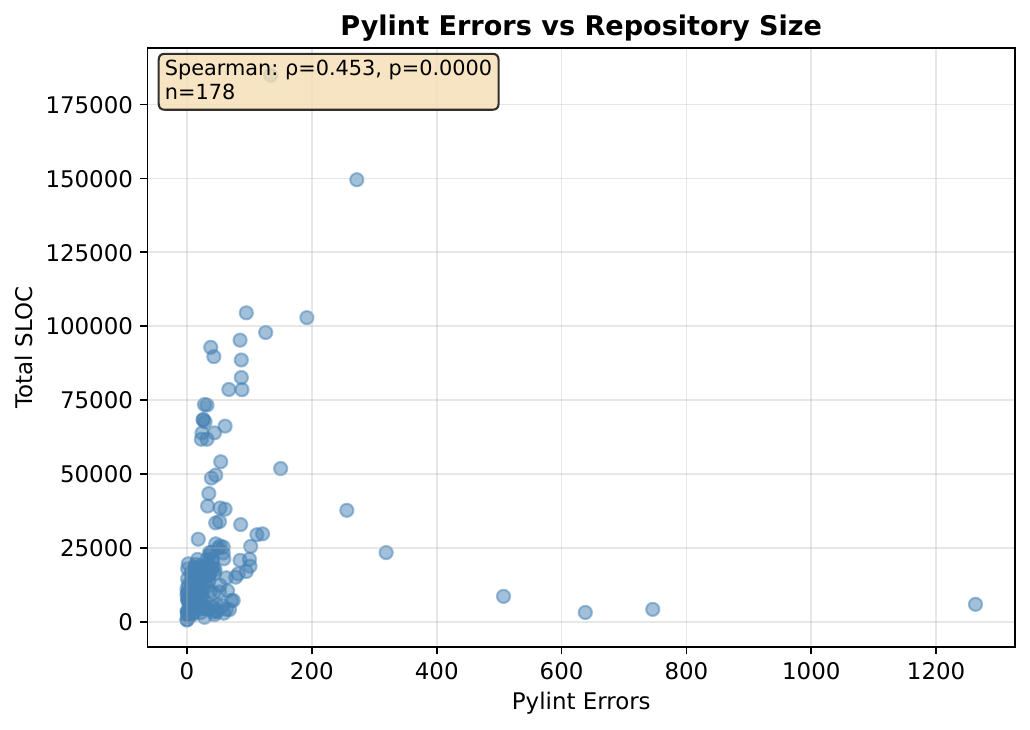}
    \caption{Relationship between errors and repository size (measured in SLOC).}
    \label{fig:errors_vs_sloc}
\end{figure}

The relationship between repository size and error counts is confirmed by Fig.~\ref{fig:errors_vs_sloc}, which shows a positive and statistically significant correlation between them. Error prevention mechanisms must therefore scale with codebase size, yet as Section~\ref{sec:sec_vulnerabilities} shows, even basic security practices are rarely adopted.

\begin{figure}
	\centering
	\includegraphics[width=.88\columnwidth]{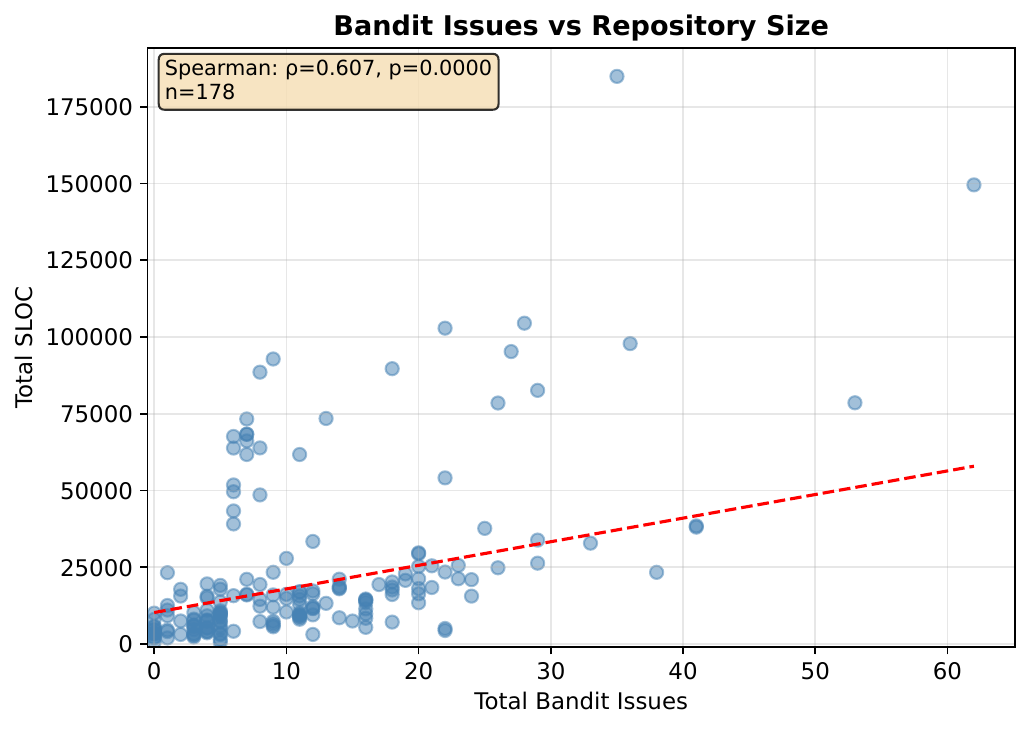}
	\caption{Correlation of repository size and number of security issues.}
	\label{fig:bandit_vs_sloc}
\end{figure}

Regarding security issues, $93.3\%$ of repositories contained at least one vulnerability, with only $12$ having zero issues. Security vulnerabilities are prevalent across Python ecosystems, with $46\%$ of PyPI packages exhibiting at least one issue~\cite{ruohonen_2021_largescale}, though the higher rate observed in AV perception repositories likely reflects research code practices that prioritize functionality over security. The number of security issues per repository ranged from $0$ to $62$, with a median of $9$ (IQR: $4\text{--}16$). Having established that code errors predictably scale with repository size, we verified whether security vulnerabilities exhibit a similar pattern. Fig.~\ref{fig:bandit_vs_sloc} shows that the number of security issues also correlates with repository size. Looking into repositories without security issues, we found that they are small, with a mean size of roughly $4.5$k SLOC. 

\begin{figure}
\centering
\includegraphics[width=.88\columnwidth]{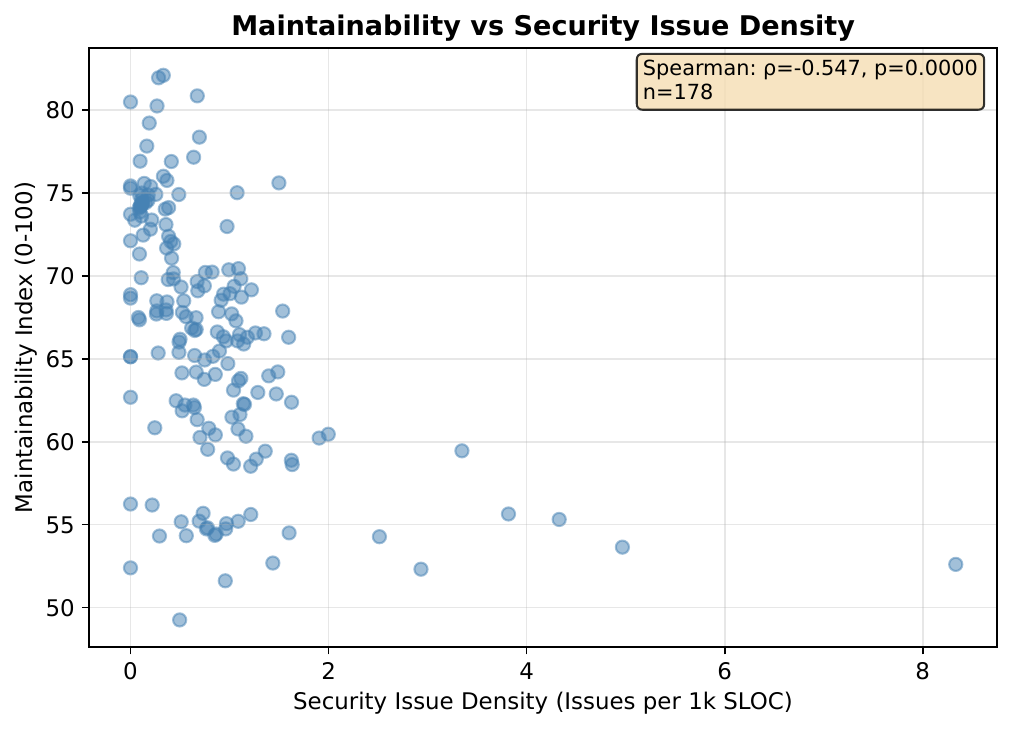}
\caption{Correlation between maintainability and security issue density.}
\label{fig:mi_vs_bandit}
\end{figure}

As shown in Fig.~\ref{fig:mi_vs_bandit}, MI correlates negatively with security issue density. This relationship also holds for error density ($\rho=-0.397$, $p<0.001$), indicating that repositories with higher maintainability tend to have fewer errors and security vulnerabilities per thousand lines of source code.

Both errors and security issues scale with codebase size, which indicates that targeted interventions are needed to prevent them and make AV perception repositories production-ready. Regarding errors specifically, Pylint identified $1,612$ total errors, $1,424$ of which were critical and $90.4\%$ of repositories having at least one critical error. Preventing these is straightforward: linters such as Pylint integrate directly into popular IDEs, and CI/CD pipelines can be configured to block commits containing critical errors.

\subsection{Security Vulnerabilities Prevention}
\label{sec:sec_vulnerabilities}

\begin{figure}
\centering
    \includegraphics[width=.985\columnwidth]{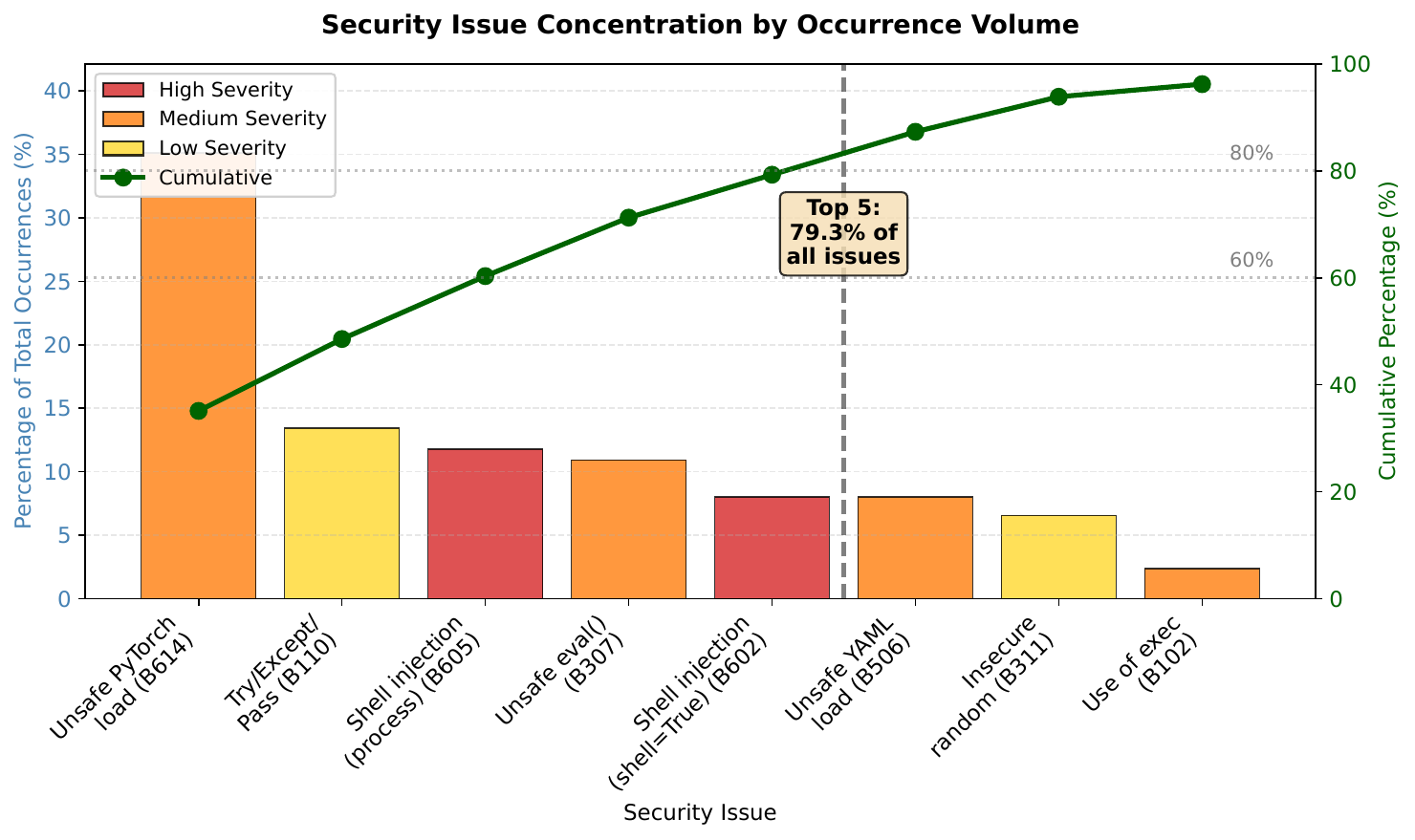}
    \caption{Concentration of security issues in our dataset. Ties broken by severity.}
    \label{fig:security_concentration}
\end{figure}

Security vulnerabilities present a less straightforward challenge. Unlike static errors, they are not always recognizable by developers without security expertise, and resolving them requires understanding the specific risk each pattern introduces, knowledge that is well-established in security practice but, as our data shows, rarely applied in AV perception research. Our analysis, however, reveals a highly actionable concentration pattern: as shown in Fig.~\ref{fig:security_concentration}, a small number of security issues account for the vast majority of occurrences.

Table~\ref{tab:bandit-issues} presents all the security issues identified by Bandit. The most common categories are serialization (affecting $83.1\%$ of repositories), input validation ($62.4\%$), shell injection ($51.7\%$), and error handling ($39.3\%$). Out of all security issues, $448$ ($23.1\%$) have low severity, $1,180$ ($58.1\%$) have medium severity, and $403$ ($19.8\%$) are of high-severity. They affect, respectively, $55.6\%$, $87.1\%$, and $51.7\%$ of repositories.

\begin{table*}[!htb]
\centering
\caption{Security issues identified by Bandit with severity classification and security categories.}
\resizebox{\textwidth}{!}{%
\begin{tabular}{@{}lllllll@{}}
\toprule
\textbf{ID} & \textbf{Severity} & \textbf{Category} & \textbf{Issue} & \textbf{Occurrences} & \textbf{\begin{tabular}[c]{@{}l@{}}Affected\\repositories\end{tabular}} & \textbf{\begin{tabular}[c]{@{}l@{}}Percentage\\of repositories\end{tabular}} \\ \midrule
B614 & MEDIUM & Serialization & Use of unsafe PyTorch load & $713$ & $144$ & $80.9$ \\
B110 & LOW & Error Handling & Try, Except, Pass & $273$ & $69$ & $38.8$ \\
B605 & HIGH & Shell Injection & Starting a process with a shell, possible injection detected & $239$ & $79$ & $44.4$ \\
B307 & MEDIUM & Input Validation & Unsafe \texttt{eval()} allowing code injection & $222$ & $106$ & $59.6$ \\
B602 & HIGH & Shell Injection & Subprocess call with \texttt{shell=True} & $163$ & $28$ & $15.7$ \\ \midrule
B506 & MEDIUM & Serialization & Use of unsafe yaml load & $163$ & $64$ & $36.0$ \\
B311 & LOW & Cryptography & Insecure random generator & $133$ & $36$ & $20.2$ \\ 
B102 & MEDIUM & Code Execution & Use of \texttt{exec} detected & $48$ & $37$ & $20.8$ \\ 
B106 & LOW & Authentication & Possible hard-coded password: `predicted' & $19$ & $7$ & $3.9$ \\
B310 & MEDIUM & Input Validation & Audit url open for permitted schemes & $12$ & $8$ & $4.5$ \\
B107 & LOW & Authentication & Possible hard-coded password: `train' & $9$ & $4$ & $2.2$ \\
B615 & MEDIUM & Dependency & Unpinned HuggingFace download & $8$ & $2$ & $1.1$ \\
B108 & MEDIUM & File System & Probable insecure usage of temp file/directory & $7$ & $6$ & $3.4$ \\
B112 & LOW & Error Handling & Try, Except, Continue detected & $5$ & $4$ & $2.2$ \\
B105 & LOW & Authentication & Possible hard-coded password: `' & $5$ & $3$ & $1.7$ \\
B405 & LOW & Input Validation & Parsing untrusted XML data with \texttt{xml.etree.ElementTree} & $2$ & $2$ & $1.1$ \\
B314 & MEDIUM & Input Validation & Parsing untrusted XML data with \texttt{xml.etree.ElementTree.parse} & $2$ & $2$ & $1.1$ \\
B408 & LOW & Input Validation & Parsing untrusted XML data with \texttt{xml.dom.minidom} & $2$ & $1$ & $0.6$ \\
B318 & MEDIUM & Input Validation & Parsing untrusted XML data with \texttt{xml.dom.minidom.parse} & $2$ & $1$ & $0.6$ \\
B104 & MEDIUM & Network & Possible binding to all interfaces & $2$ & $2$ & $1.1$ \\
B324 & HIGH & Cryptography & Use of weak SHA1 hash for security & $1$ & $1$ & $0.6$ \\
B113 & MEDIUM & Network & Call to requests without timeout & $1$ & $1$ & $0.6$ \\ \bottomrule \\
\end{tabular}%
}
\label{tab:bandit-issues}
\end{table*}

We therefore investigated samples of these issues and developed prevention guidelines for the five most common patterns.

\subsubsection{B614: Unsafe PyTorch Model Loading}

In AV perception pipelines, model checkpoints are frequently shared between research groups and loaded during inference. These are loaded through PyTorch's \texttt{torch.load()}, which uses Python's pickle module for deserialization. This module is not only able to store data, but can also store executable instructions for reconstructing Python objects, meaning that loading a malicious checkpoint file could trigger arbitrary code execution before any user code runs. As such, a compromised checkpoint could silently alter detection thresholds or classification boundaries, causing the perception system to systematically miss obstacles or misclassify road users without any observable error. This vulnerability affects the majority of repositories in our dataset, with only $34$ not being affected.

\textbf{Vulnerable Pattern:}
\nopagebreak
\begin{badcode}
\begin{lstlisting}
# Allows arbitrary code execution
checkpoint = torch.load('model.pth')
model.load_state_dict(torch.load('weights.pth'))
\end{lstlisting}
\end{badcode}

\textbf{Secure Implementation:}
\nopagebreak
\begin{goodcode}
\begin{lstlisting}
# Only loads tensor data, blocks arbitrary code
checkpoint = torch.load('model.pth', weights_only=True)
model.load_state_dict(torch.load('weights.pth', weights_only=True))
\end{lstlisting}
\end{goodcode}

\subsubsection{B110: Silent Error Suppression}

Despite being classified by Bandit as low-severity, this issue is particularly dangerous for safety-critical AV perception. By silently suppressing exceptions, critical errors would be hidden and issues such as a failed LiDAR point cloud read, a corrupted camera frame, or a sensor synchronization error would go undetected. This could then cause the perception system to operate on faulty data while reporting nominal status to downstream modules.

\textbf{Vulnerable Pattern:}
\nopagebreak
\begin{badcode}
\begin{lstlisting}
try:
    model.load_state_dict(checkpoint)
except:
    pass  # Silently fails - hard to debug
\end{lstlisting}
\end{badcode}

\textbf{Secure Implementation:}
\nopagebreak
\begin{goodcode}
\begin{lstlisting}
try:
    model.load_state_dict(checkpoint)
except Exception as e:
    logger.error(f"Failed to load checkpoint: {e}")
    raise  # Or handle appropriately
\end{lstlisting}
\end{goodcode}

\subsubsection{B605/B602: Shell Injection}

Perception systems often invoke external preprocessing tools for point cloud transformation, sensor calibration, or data format conversion. Using shell-based subprocess calls in these pipelines creates injection vectors that could be exploited to manipulate sensor data before it reaches the detection model, undermining the integrity of the entire perception chain.

B605 and B602 each appear among the five most common security issues but represent the same underlying vulnerability through different mechanisms: B605 is triggered by \texttt{os.system()} and related functions that implicitly invoke a shell, while B602 is triggered by \texttt{subprocess} calls with \texttt{shell=True} explicitly set. In both cases, unsanitized string input is passed to a shell interpreter, and the remediation strategy is identical.

\textbf{Vulnerable Pattern:}
\nopagebreak
\begin{badcode}
\begin{lstlisting}
import subprocess
cmd = f"rm {user_input}"  
# user_input could be "; rm -rf /"
subprocess.call(cmd, shell=True)
\end{lstlisting}
\end{badcode}

\textbf{Secure Implementation:}
\nopagebreak
\begin{goodcode}
\begin{lstlisting}
import subprocess
# List syntax prevents injection
subprocess.call(['rm', user_input])
\end{lstlisting}
\end{goodcode}

\subsubsection{B307: Unsafe Use of eval()}

In perception systems, \texttt{eval()} is sometimes used to dynamically instantiate model architectures or parse configuration strings specifying network layers. In a deployment context, this allows an attacker who can modify a configuration file or model definition to execute arbitrary code within the perception process, potentially disabling object detection or injecting false detections at runtime.

\textbf{Vulnerable Pattern:}
\nopagebreak
\begin{badcode}
\begin{lstlisting}
# Executes arbitrary code from string
eval(args.task)(log_dicts, args)
result = eval(user_input)
\end{lstlisting}
\end{badcode}

\textbf{Secure Implementation:}
\nopagebreak
\begin{goodcode}
\begin{lstlisting}
# Use explicit function mapping instead of eval
TASKS = {'train': train_task, 'validate': validate_task}
task_fn = TASKS.get(args.task)
if task_fn: task_fn(log_dicts, args)
\end{lstlisting}
\end{goodcode}

\subsection{Development Practices}

Having established actionable guidelines for the most prevalent security issues, we next examined whether development practices offer a complementary path to improving code quality. Despite collecting several statistics from GitHub for each repository, most showed no correlation with errors, security issues, or maintainability. Most notably, repository popularity is not indicative of code quality, with stars and errors showing no significant correlation ($\rho=0.04$, $p=0.56$). This means that despite popular repositories often being favored by practitioners, they are no more production-ready than their less popular counterparts. The one development practice that did correlate meaningfully with quality was CI/CD adoption.

\begin{figure}
\centering
    \includegraphics[width=\columnwidth]{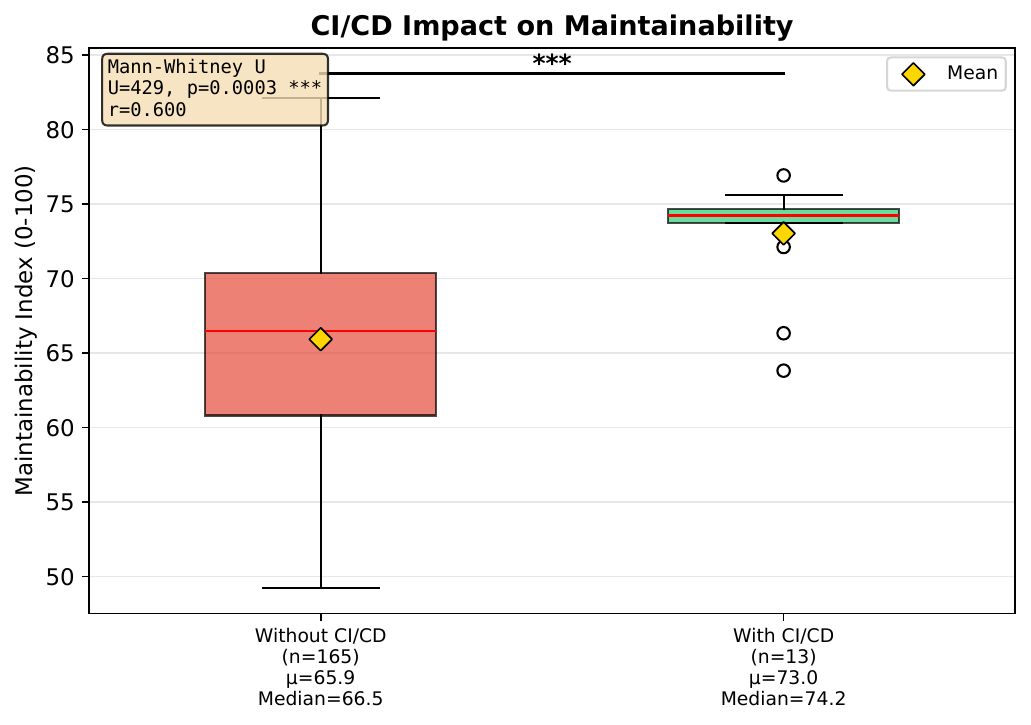}
    \caption{Repositories that adopt CI/CD have a significantly higher mean maintainability index.}
    \label{fig:cicd_vs_maintainability}
\end{figure}

Although CI/CD practices are used only by $7.3\%$ of repositories in our dataset, those that adopt CI/CD show higher mean MI than those that do not, as shown in Fig.~\ref{fig:cicd_vs_maintainability}. This is noteworthy given that MI correlates with lower error and security issue densities (Section~\ref{sec:res_production_readiness}). However, repositories with CI/CD also tend to have more contributors (mean $34.4$ vs $2.3$), making it unclear whether CI/CD directly impacts MI or if the larger number of contributors is solely responsible.

We investigated the matter further and found that repositories that adopt CI/CD have higher mean MI than their counterparts that do not, regardless of team size. Across all team sizes, repositories with CI/CD consistently achieved higher mean MI ($69.2$ to $74.2$) than those without ($65.4$ to $69.9$), with the largest difference observed in medium-sized teams ($3-5$ contributors). According to established MI thresholds~\cite{coleman_1994_using}, these values fall within the moderately maintainable range ($65 \leq MI < 85$), with repositories without CI/CD approaching the low maintainability threshold ($MI < 65$). This suggests that CI/CD adoption correlates with higher MI regardless of team size. However, the low sample sizes make this comparison not statistically significant, and further studies are required to validate this relationship. 

Since only $7.3\%$ of the repositories in our dataset adopt CI/CD, these findings suggest that broader adoption represents one of the most actionable paths towards closing the research-production gap in AV perception code.

\section{Limitations}
\label{sec:limitations}

Despite building our dataset from the two largest AV perception leaderboards, using only two leaderboards as the source for our data posed an external threat to validity. We aimed to alleviate this threat by the large number of repositories ($178$) analyzed. Selection bias may also have affected our findings, as we only analyzed repositories published on leaderboards. Published AV perception codebases may be different from unpublished ones in terms of maturity and quality.

Internal threats to validity include the use of static analysis tools to assess repository quality, particularly Pylint's error detection, which is subject to both false positives and false negatives. There is also the matter of construct validity, which concerns whether our chosen metrics truly measure code quality. While errors, security issues, and MI are widely accepted metrics for code quality, they may not capture all aspects of software quality relevant to production deployment.

\section{Conclusion}
\label{sec:conclusions}

We have conducted the first large-scale software quality analysis of AV perception repositories, evaluating $178$ repositories from the largest AV 3D Object Detection leaderboards: KITTI and NuScenes. Our main findings were:

\begin{itemize}
	\item Code quality is particularly low, with only $2.8\%$ of the repositories being error-free, and only $6.7\%$ having no security vulnerabilities.
	\item Considering the absence of critical errors and high-severity security issues, only $7.3\%$ of repositories were deemed production-ready.
	\item We found a strong concentration of security issues, with five issues accounting for almost $80\%$ of occurrences.
	\item Although only $7.3\%$ of repositories analyzed adopt CI/CD practices, we found that CI/CD adoption correlates with more maintainable code.
\end{itemize}

These findings are directly relevant to AV deployment, as safety standards such as ISO 26262 and SOTIF (ISO 21448) require systematic use of static analysis and coding guidelines. The prevalence of security vulnerabilities and near-absence of CI/CD adoption in research perception repositories indicates that significant remediation is needed before they can be integrated into safety-certified AV systems.

Future work could expand our analysis beyond 3D Object Detection, including Semantic Segmentation, Tracking, and Collaborative Perception. Additionally, dynamic analysis through code execution could provide deeper insights into runtime behavior and model quality.
    
\bibliographystyle{IEEEtran}
\bibliography{root} 

@String{Computer = "{IEEE} Computer" }

@article{caesar_2019_nuscenes,
   author = {Holger Caesar and Varun Bankiti and Alex H Lang and Sourabh Vora and Venice Erin Liong and Qiang Xu and Anush Krishnan and Yu Pan and Giancarlo Baldan and Oscar Beijbom and Aptiv Company},
   isbn = {978-1-7281-7168-5},
   issn = {10636919},
   journal = {2020 IEEE/CVF Conference on Computer Vision and Pattern Recognition (CVPR)},
   month = {6},
   pages = {11618-11628},
   title = {nuScenes: A multimodal dataset for autonomous driving},
   year = {2020},
}

@article{geiger_2013_kitti,
   author = {Andreas Geiger and Philip Lenz and Christoph Stiller and Raquel Urtasun},
   issue = {11},
   journal = {International Journal of Robotics Research},
   pages = {1231-1237},
   title = {Vision meets Robotics: The KITTI Dataset},
   volume = {32},
   url = {http://www.cvlibs.net/datasets/kitti.},
   year = {2013},
}

@software{pylint_2025,
  title = {Pylint},
  subtitle = {It's Not Just a Linter That Annoys You!},
  author = {{Pylint Development Team}},
  year = {2025},
  url = {https://pylint.readthedocs.io/},
  howpublished = {\url{https://github.com/pylint-dev/pylint}},
  license = {GPL-2.0},
  note = {Python static code analysis tool that checks for errors, enforces coding standards, and looks for code smells}
}

@software{bandit_2025,
	author = {{PyCQA}},
	title = {Bandit: Security Linter for Python},
	year = {2025},
	howpublished = {\url{https://github.com/PyCQA/bandit}},
	note = {Accessed: Nov. 2025}
}

@misc{radon_2025,
	author = {M. Lacchia},
	title = {Radon: Code Metrics for Python},
	year = {2025},
	howpublished = {\url{https://github.com/rubik/radon}},
	note = {Accessed: Nov. 2025}
}

@techreport{MISRA2012,
  author = {{MISRA}},
  title = {{MISRA C:2012 Guidelines for the use of the C language in critical systems}},
  institution = {Motor Industry Software Reliability Association},
  year = {2012},
  edition = {3rd},
  note = {{ISBN} 978-1-906400-10-1}
}

@ARTICLE{coleman_1994_using,
  author={Coleman, D. and Ash, D. and Lowther, B. and Oman, P.},
  journal={Computer}, 
  title={Using metrics to evaluate software system maintainability}, 
  year={1994},
  volume={27},
  number={8},
  pages={44-49},
  doi={10.1109/2.303623}}

@techreport{salay_2018_analysis,
  title={An analysis of {ISO} 26262: Machine learning and safety in automotive software},
  author={Salay, Rick and Queiroz, Rodrigo and Czarnecki, Krzysztof},
  year={2018},
  institution={SAE Technical Paper}
}

@article{koopman_2016_challenges,
  title={Challenges in autonomous vehicle testing and validation},
  author={Koopman, Philip and Wagner, Michael},
  journal={SAE International Journal of Transportation Safety},
  volume={4},
  number={1},
  pages={15--24},
  year={2016},
  publisher={JSTOR}
}

@inproceedings{tabani_2019_assessing,
  title={Assessing the adherence of an industrial autonomous driving framework to {ISO} 26262 software guidelines},
  author={Tabani, Hamid and Kosmidis, Leonidas and Abella, Jaume and Cazorla, Francisco J and Bernat, Guillem},
  booktitle={Proceedings of the 56th Annual Design Automation Conference 2019},
  pages={1--6},
  year={2019}
}

@article{cheng_2025_unveiling,
  title={Unveiling security weaknesses in autonomous driving systems: An in-depth empirical study},
  author={Cheng, Wenyuan and Li, Zengyang and Liang, Peng and Mo, Ran and Liu, Hui},
  journal={Information and Software Technology},
  volume={182},
  pages={107709},
  year={2025},
  publisher={Elsevier}
}

@article{kochanthara_2024_safety,
  title={Safety of perception systems for automated driving: A case study on apollo},
  author={Kochanthara, Sangeeth and Singh, Tajinder and Forrai, Alexandru and Cleophas, Loek},
  journal={ACM Transactions on Software Engineering and Methodology},
  volume={33},
  number={3},
  pages={1--28},
  year={2024},
  publisher={ACM New York, NY}
}

@techreport{iso26262,
  author      = {{ISO}},
  title       = {{ISO} 26262: Road Vehicles --- Functional Safety},
  institution = {International Organization for Standardization},
  address     = {Geneva, Switzerland},
  year        = {2018}
}

@techreport{iso21448,
  author      = {{ISO}},
  title       = {{ISO} 21448: Road Vehicles --- Safety of the Intended Functionality ({SOTIF})},
  institution = {International Organization for Standardization},
  address     = {Geneva, Switzerland},
  year        = {2022}
}

@INPROCEEDINGS{breck_2017_mltest,
  author={Breck, Eric and Cai, Shanqing and Nielsen, Eric and Salib, Michael and Sculley, D.},
  booktitle={2017 IEEE International Conference on Big Data (Big Data)}, 
  title={The {ML} test score: A rubric for {ML} production readiness and technical debt reduction}, 
  year={2017},
  volume={},
  number={},
  pages={1123-1132},
  doi={10.1109/BigData.2017.8258038}}

@INPROCEEDINGS {ruohonen_2021_largescale,
author = { Ruohonen, Jukka and Hjerppe, Kalle and Rindell, Kalle },
booktitle = { 2021 18th International Conference on Privacy, Security and Trust (PST) },
title = {{A Large-Scale Security-Oriented Static Analysis of Python Packages in PyPI}},
year = {2021},
volume = {},
ISSN = {},
pages = {1-10},
doi = {10.1109/PST52912.2021.9647791},
url = {https://doi.ieeecomputersociety.org/10.1109/PST52912.2021.9647791},
publisher = {IEEE Computer Society},
address = {Los Alamitos, CA, USA},
month =Dec}
	
\end{document}